\def\year{2019}\relax
\documentclass[letterpaper]{article} 
\pdfoutput=1
\usepackage{aaai19}  
\usepackage{times}  
\usepackage{helvet}  
\usepackage{courier}  
\usepackage{hyperref}       
\usepackage{url}            
\usepackage{booktabs}       
\usepackage{amsfonts}       
\usepackage{nicefrac}       
\usepackage{microtype}      
\usepackage{subfigure}
\usepackage{makecell}
\usepackage{multirow}
\usepackage{enumitem}
\usepackage{amsmath}
\usepackage{enumerate}
\usepackage{xcolor}
\usepackage{bm}
\usepackage{graphicx}
\usepackage{booktabs}
\usepackage{subfigure}
\usepackage{hyperref}
\usepackage{color}
\usepackage{chngpage}
\usepackage{import}
\usepackage{amssymb,amsfonts}
\usepackage{algorithmic}
\usepackage{textcomp}
\usepackage{flushend}
\usepackage{hhline}
\usepackage{multirow}
\usepackage{array}
\usepackage{gensymb}
\usepackage{epstopdf}
\usepackage{nameref}
\usepackage{tabularx}
\graphicspath{{./figs/}}
\newcolumntype{"}{!{\vrule width 1pt}}
\newcolumntype{L}[1]{>{\raggedright\let\newline\\\arraybackslash\hspace{0pt}}m{#1}}
\newcolumntype{C}[1]{>{\centering\let\newline\\\arraybackslash\hspace{0pt}}m{#1}}
\newcolumntype{R}[1]{>{\raggedleft\let\newline\\\arraybackslash\hspace{0pt}}m{#1}}

\def\allinone{all_in_one} 
\begin{document}
\title{GaitSet: Regarding Gait as a Set for Cross-View Gait Recognition
\thanks{This work is supported in part by National Natural Science Foundation of
China (NSFC) (Grant No. 61673118) and in part by Shanghai Pujiang Program
(Grant No. 16PJD009).}
}
\author{Hanqing Chao$^1$$^{\dagger}$, Yiwei He$^1$\thanks{H.C. and Y.H. are co-first authors.}, Junping Zhang$^1$\thanks{Corresponding author.}, JianFeng Feng$^2$\\
$^1$Shanghai Key Laboratory of Intelligent
Information Processing, School of Computer Science\\
$^2$Institute of Science and Technology for Brain-inspired Intelligence\\
Fudan University, Shanghai 200433, China\\
\{hqchao16, heyw15, jpzhang, jffeng\}@fudan.edu.cn\\
}
\maketitle

\begin{abstract}
As a unique biometric feature that can be recognized at a distance, gait has broad applications in crime prevention, forensic identification and social security. To portray a gait, existing gait recognition methods utilize either a gait template, where temporal information is hard to preserve, or a gait sequence, which must keep unnecessary sequential constraints and thus loses the flexibility of gait recognition. In this paper we present a novel perspective, where a gait is regarded as a \emph{set} consisting of independent frames. We propose a new network named \emph{GaitSet} to learn identity information from the \emph{set}. Based on the \emph{set} perspective, our method is \emph{immune to permutation} of frames, and can naturally \emph{integrate frames from different videos} which have been filmed under different scenarios, such as diverse viewing angles, different clothes/carrying conditions. Experiments show that under normal walking conditions, our single-model method achieves an average rank-1 accuracy of 95.0\% on the CASIA-B gait dataset and an 87.1\% accuracy on the OU-MVLP gait dataset. These results represent new state-of-the-art recognition accuracy. On various complex scenarios, our model exhibits a significant level of robustness. It achieves accuracies of 87.2\% and 70.4\% on CASIA-B under bag-carrying and coat-wearing walking conditions, respectively. These outperform the existing best methods by a large margin. The method presented can also achieve a satisfactory accuracy with a small number of frames in a test sample, e.g., 82.5\% on CASIA-B with only 7 frames. The source code has been released at \url{https://github.com/AbnerHqC/GaitSet}.
\end{abstract}

\ifx\allinone\undefined
\input{../preamble}
\begin{document}
\fi

\section{Introduction}
Unlike other biometrics such as face, fingerprint and iris,  gait is a unique biometric feature that can be recognized at a distance without the cooperation of subjects and intrusion to them. Therefore, it has broad applications in crime prevention, forensic identification and social security.

However, gait recognition suffers from exterior factors such as the subject's walking speed, dressing and carrying condition, and the camera's viewpoint and frame rate. There are two main ways to identify gait in literature, i.e., regarding gait as an image and regarding gait as a video sequence. The first category compresses all gait silhouettes into one image, or gait template for gait recognition~\cite{he2019multi,takemura2017input,wu2017comprehensive,hu2013view}. Simple and easy to implement, gait template easily loses temporal and fine-grained spatial information. Differently, the second category extracts features directly from the original gait silhouette sequences in recent years~\cite{liao2017pose,takemura2018multi}. However, these methods are vulnerable to 
exterior factors. Further, deep neural networks like 3D-CNN for extracting sequential information are harder to train than those using a single template like Gait Energy Image~(GEI)~\cite{han2006individual}.

\begin{figure}[t]
\centering
\includegraphics[width=1\linewidth, clip=true, trim=70 170 70 170]{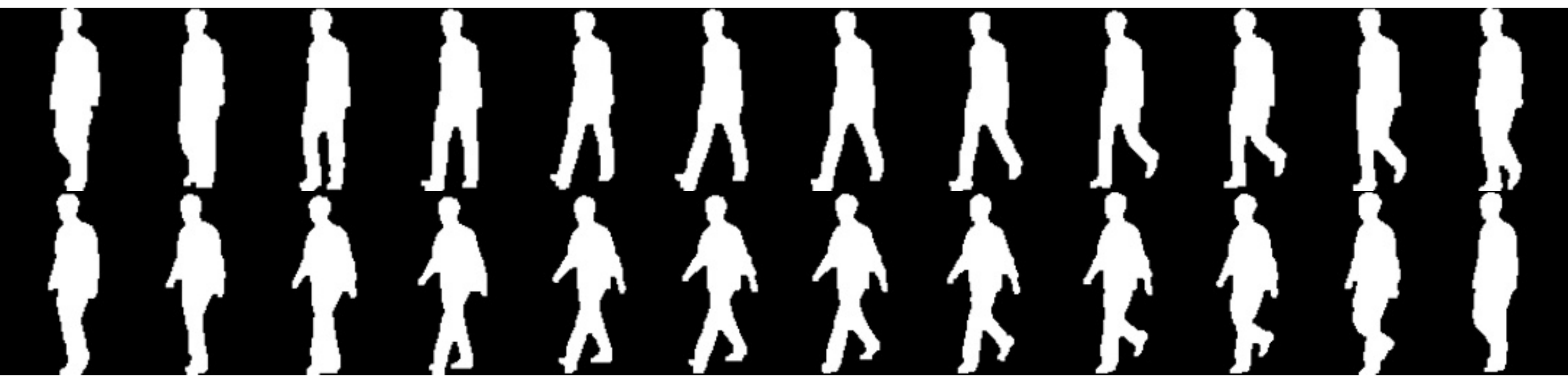}
\caption{ From top-left to bottom-right are silhouettes of a
completed period of a subject in \textbf{CASIA-B} gait dataset.}
\label{fig:gaitseq}
\end{figure}

To solve these problems, we present a novel perspective which regards gait as a set of gait silhouettes. As a periodic motion, gait can be represented by a single period. In a silhouette sequence containing one gait period, it was observed that the silhouette in each \textbf{\emph{position}} has unique appearance, as shown in Fig.~\ref{fig:gaitseq}. Even if these silhouettes are shuffled, it is not difficult to rearrange them into correct order only by observing the appearance of them. Thus, we assume the appearance of a silhouette has contained its \textbf{\emph{position}} information. With this assumption, order information of gait sequence is not necessary and we can directly regard gait as a set to extract temporal information. We propose an end-to-end deep learning model called GaitSet whose scheme is shown in Fig.~\ref{fig:pipeline}. The input of our model is a set of gait silhouettes. First, a CNN is used to extract frame-level features from each silhouette independently. Second, an operation called Set Pooling is used to aggregate frame-level features into a single \emph{set-level} feature. Since this operation is applied on high-level feature maps instead of the original silhouettes, it can preserve spatial and temporal information better than gait template. This will be justified by the experiment in Sec.~\ref{sec:abl}. Third, a structure called Horizontal Pyramid Mapping is used to map the set-level feature into a more discriminative space to obtain the final representation.
The superiorities of the proposed method are summarized as follows:
\begin{itemize}
\item \textbf{Flexible} Our model is pretty flexible since there are no any constraints on the input of our model except the size of the silhouette. It means that the input set can contain any number of non-consecutive silhouettes filmed under different viewpoints with different walking conditions. 
Related experiments are shown in Sec.~\ref{sec:pra}
\item \textbf{Fast} Our model directly learns the representation of gait instead of measuring the similarity between a pair of gait templates or sequences. Thus, the representation of each sample needs to be calculated only once, then the recognition can be completed by calculating the Euclidean distance between representations of different samples.
\item \textbf{Effective} Our model greatly improves the performance on the CASIA-B~\cite{yu2006framework} and the OU-MVLP~\cite{Takemura2018} datasets, showing its strong robustness to view and walking condition variations and high generalization ability to large datasets.
\end{itemize}


\ifx\allinone\undefined
\newpage
\bibliographystyle{aaai}
\bibliography{../ref}
\end{document}
\fi
\ifx\allinone\undefined
\input{../preamble}
\begin{document}
\fi

\section{Related Work}
\label{sec:rw}
In this section, we will give a brief survey on gait recognition and set-based deep learning methods. 
\subsection{Gait Recognition}
Gait recognition can be grouped into template-based and sequence-based categories. Approaches in the former category first obtain human silhouettes of each frame by background subtraction. Second, they generate a gait template by rendering pixel level operators on the aligned silhouettes~\cite{han2006individual,wang2012human}. Third, they extract the representation of the gait by machine learning approaches such as Canonical Correlation Analysis~(CCA)~\cite{xing2016complete}, Linear Discriminant Analysis~(LDA)~\cite{bashir2010gait} and deep learning~\cite{shiraga2016geinet}. Fourth, they measure the similarity between pairs of representations by Euclidean distance or some metric learning approaches~\cite{wu2017comprehensive,takemura2017input}. Finally, they assign a label to the template by some classifier, e.g., nearest neighbor classifier.

Previous works generally divides this pipeline into two parts, template generation and matching. The goal of generation is to compress gait information into a single image, e.g., Gait Energy Image~(GEI)~~\cite{han2006individual} and Chrono-Gait Image~(CGI)~\cite{wang2012human}. In template matching approaches, View Transformation Model~(VTM) learns a projection between different views~\cite{makihara2006gait}. \cite{hu2013view} proposed View-invariant Discriminative Projection~(ViDP) to project the templates into a latent space to learn a view-invariance representation. Recently, as deep learning performs well on various generation tasks, it has been employed on gait recognition task~\cite{yu2017gaitgan,he2019multi,takemura2017input,shiraga2016geinet,yu2017invariant,wu2017comprehensive}.

As the second category, video-based approaches directly take a sequence of silhouettes as input. Based on the way of extracting temporal information,  they can be classified into LSTM-based approaches~\cite{liao2017pose} and 3D CNN-based approaches~\cite{takemura2018multi,wu2017comprehensive}. The advantages of these approaches are that \textbf{1)} focusing on each silhouette, they can obtain more comprehensive spatial information. \textbf{2)} They can gather more temporal information because specialized structures are utilized to extract sequential information. However, The price to pay for these advantages is high computational cost.

\subsection{Deep learning on Unordered set}
Most works in deep learning focus on regular input representations like sequence and images. The concept of unordered set is first introduced into computer vision by \cite{qi2017pointnet}~(PointNet) to tackle point cloud tasks. Using unordered set, PointNet can avoid the noise and the extension of data caused by quantization, and obtain a high performance. Since then, set-based methods have been wildly used in point cloud field~\cite{wang2018dynamic,zhou2017voxelnet,qi2017pointnet++}. Recently, such methods are introduced into computer vision domains like content recommendation~\cite{hamilton2017inductive} and image captioning~\cite{krause2017hierarchical} to aggregate features in a form of a set. \cite{zaheer2017deep} further formalized the deep learning tasks defined on sets and characterizes the permutation invariant functions. To the best of our knowledge, it has not been employed in gait recognition domain up to now.

\ifx\allinone\undefined
\newpage
\bibliographystyle{aaai}
\bibliography{../ref}
\end{document}
\fi
\ifx\allinone\undefined
\input{../preamble}
\begin{document}
\fi

\section{GaitSet}
\label{sec:method}

\begin{figure*}[t]
\centering
\includegraphics[width=1\linewidth, clip=true, trim=0 141 0 142]{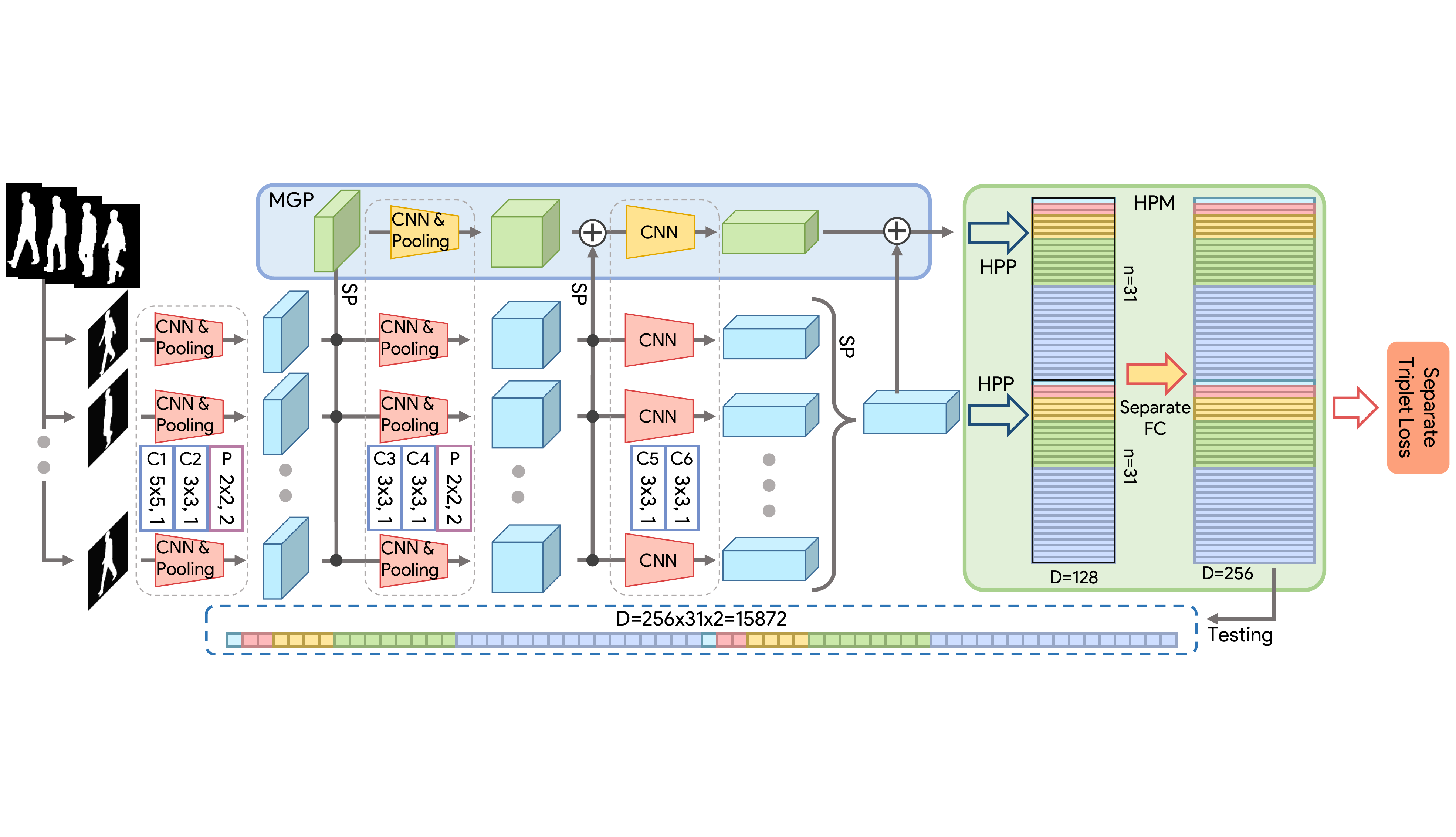}
\caption{The framework of GaitSet. 
'SP' represents Set Pooling. Trapezoids represent convolution and pooling blocks and those in the same column have the same configurations which are shown by rectangles with capital letters. Note that although blocks in MGP have same configurations with those in the main pipeline, parameters are only shared across blocks in the main pipeline but not with those in MGP. HPP represents horizontal pyramid pooling~\protect\cite{fu2018horizontal}.}
\label{fig:pipeline}
\end{figure*}

In this section, we describe our method for learning discriminative information from a set of gait silhouettes. The overall pipeline is illustrated in Fig.~\ref{fig:pipeline}.

\subsection{Problem Formulation}
We begin with formulating our concept of regarding gait as a set. Given a dataset of $N$ people with identities $y_i, i\in{1,2, ..., N}$, we assume the gait silhouettes of a certain person subject to a distribution $\mathcal{P}_i$ which is only related to its identity. Therefore, all silhouettes in one or more sequences of a person can be regarded as a set of $n$ silhouettes $\mathcal{X}_i=\{x_i^j|j=1,2,...,n\}$, where $x_i^j\sim\mathcal{P}_i$. 

Under this assumption, we tackle the gait recognition task through 3 steps, formulated as
\begin{equation}
f_i=H(G(F(\mathcal{X}_i)))
\end{equation}
where $F$ is a convolutional network aims to extract frame-level features from each gait silhouette. The function $G$ is a permutation invariant function used to map a set of frame-level feature to a set-level feature~\cite{zaheer2017deep}. It is implemented by an operation called Set Pooling~(SP) which will be introduced in Sec.~\ref{sec:SP}. The function $H$ is used to learn the discriminative representation of $\mathcal{P}_i$ from the set-level feature. This function is implemented by a structure called Horizontal Pyramid Mapping~(HMP) which will be discussed in Sec.~\ref{sec:HPM}. The input $\mathcal{X}_i$ is a tensor with four dimensions, i.e. set dimension, image channel dimension, image hight dimension, and image width dimension.

\subsection{Set Pooling}\label{sec:SP}
The goal of Set Pooling~(SP) is to aggregate gait information of elements in a set, formulated as $z=G(V)$, 
where $z$ denotes the set-level feature and $V=\{v^j|j=1,2,...,n\}$ denotes the frame-level features. There are two constraints in this operation. First, to take set as an input, it should be a permutation invariant function which is formulated as:
\begin{equation}
\label{eq:sp}
G(\{v^j|j=1,2,...,n\}) = G(\{v^{\pi(j)}|j=1,2,...,n\})
\end{equation}
where $\pi$ is any permutation~\cite{zaheer2017deep}. Second, since in real-life scenario the number of a person's gait silhouettes can be arbitrary, the function $G$ should be able to take a \emph{set} with arbitrary cardinality. Next, we describe several instantiations of $G$. It will be shown in the experiments that although different instantiations of SP do have sort of influence on the performances, they do not differ greatly and all of them exceed GEI-based methods by a large margin.

\textbf{Statistical Functions}~~
To meet the requirement of invariant constraint in Equ.~\ref{eq:sp}, a natural choice of SP is to apply statistical functions on the set dimension. Considering the representativeness and the computational cost, we studied three statistical functions: $\mathrm{max}(\cdot)$, $\mathrm{mean}(\cdot)$ and $\mathrm{median}(\cdot)$. The comparison will be shown in Sec.~\ref{sec:abl}.

\textbf{Joint Functions}~~
We also studied two ways to join $3$ statistical functions mentioned above:
\begin{align}
\label{eq:sum}
G(\cdot)&=\mathrm{max}(\cdot)+\mathrm{mean}(\cdot)+\mathrm{median}(\cdot) \\
\label{eq:cat}
G(\cdot)&=\mathrm{1\_1C}(\mathrm{cat}(\mathrm{max}(\cdot),
 \mathrm{mean}(\cdot), \mathrm{median}(\cdot)))
\end{align}
where $\mathrm{cat}$ means concatenate on the channel dimension, $1\_1C$ means $1\times1$ convolutional layer, and $\mathrm{max}$, $\mathrm{mean}$ and $\mathrm{median}$ are applied on set dimension. Equ.~\ref{eq:cat} is an enhanced version of Equ.~\ref{eq:sum} where the $1\times1$ convolutional layer can learn a proper weight to combine information extracted by different statistical functions.

\begin{figure}[htbp]
\centering
\includegraphics[width=1\linewidth, clip=true, trim=240 200 240 200]{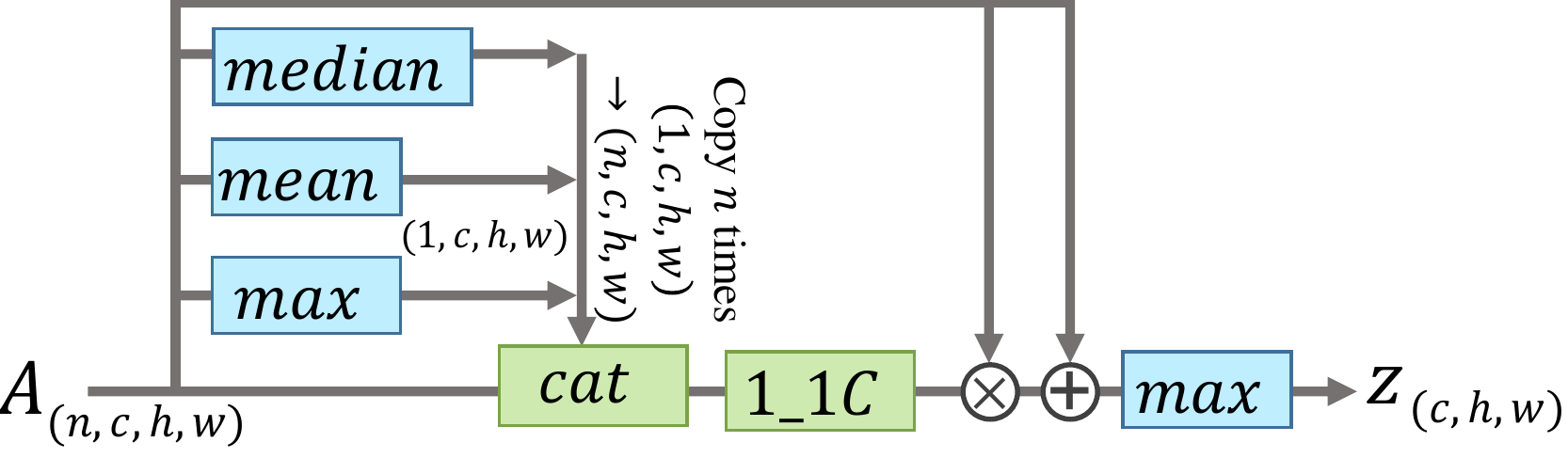}
\caption{The structure of Set Pooling~(SP) using attention. $1\_1C$ and $cat$ represents $1\times1$ convolutional layer and concatenate respectively. The multiplication and the addition are both pointwise.}
\label{fig:attention}
\end{figure}

\textbf{Attention}~~
Since visual attention was successfully applied in lots of tasks~\cite{wang2018non,xu2015show,li2018harmonious}, we use it to improve the performance of SP. Its structure is shown in Fig.~\ref{fig:attention}. The main idea is to utilize the global information to learn an element-wise attention map for each frame-level feature map to refine it.
Global information is first collected by the statistical functions in the left. Then it is fed into a $1\times1$ convolutional layer along with the original feature map to calculate an attention for the refinement. The final set-level feature $z$ will be extracted by employing \emph{MAX} on the set of the refined frame-level feature maps. The residual structure can accelerate and stabilize the convergence.

\subsection{Horizontal Pyramid Mapping}\label{sec:HPM}

\begin{figure}[htbp]
\centering
\includegraphics[width=1\linewidth, clip=true, trim=70 80 70 100]{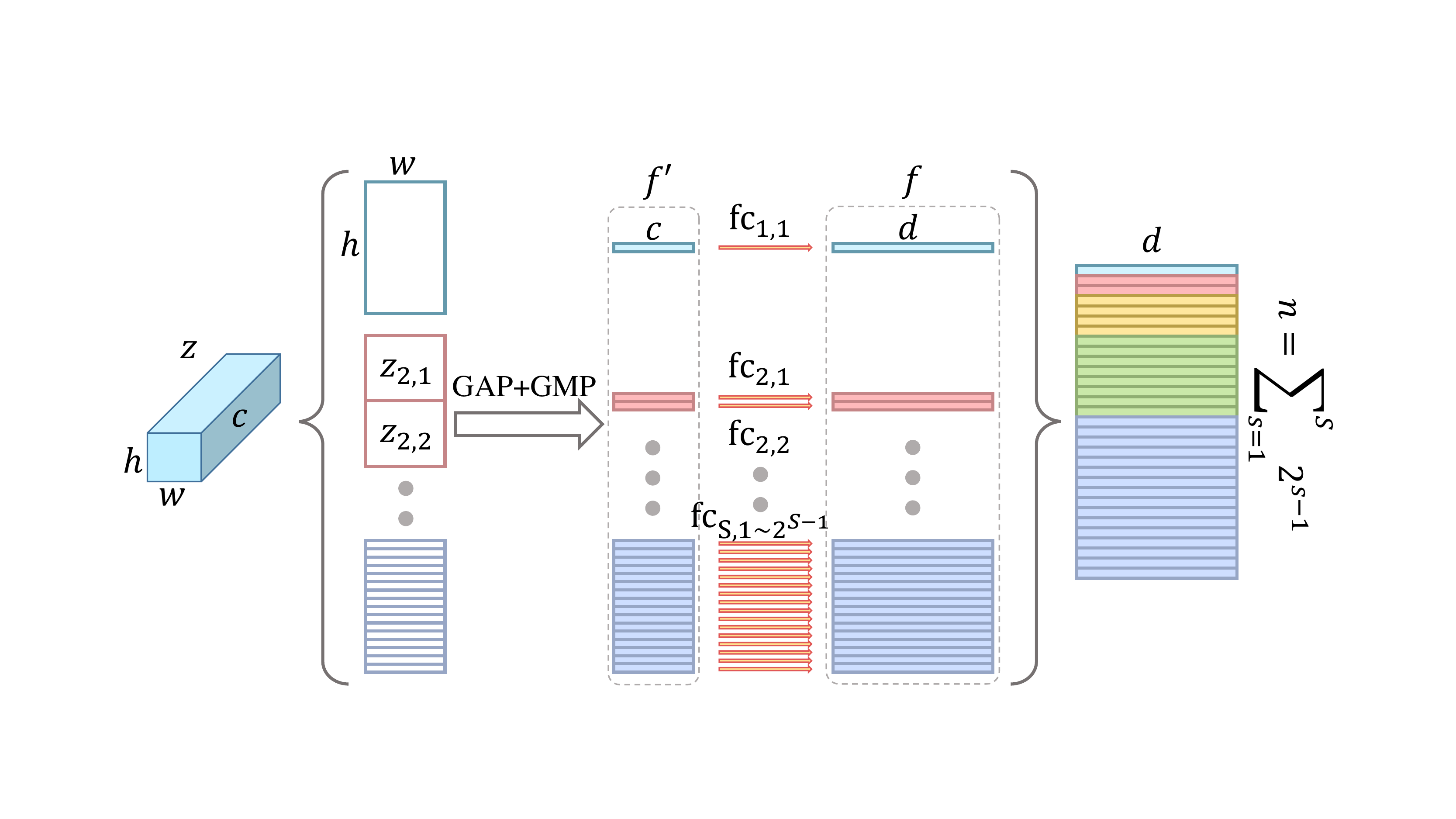}
\caption{The structure of Horizontal Pyramid Mapping.}
\label{fig:hpm}
\end{figure}

In literature, splitting feature map into strips is commonly used in person re-identification task~\cite{wang2018learning,fu2018horizontal}. The images are cropped and resized into uniform size according to pedestrian size whereas the discriminative parts vary from image to image. \cite{fu2018horizontal} proposed Horizontal Pyramid Pooling (HPP) to deal with it. HPP has 4 scales and thus can help the deep network focus on features with different sizes to gather both local and global information. We improve HPP to make it adapt better for gait recognition task. Instead of applying a $1\times1$ convolutional layer after the pooling, we use independent fully connect layers~(FC) for each pooled feature to map it into the discriminative space, as shown in Fig.~\ref{fig:hpm}. We call it Horizontal Pyramid Mapping (HPM).

Specifically, HPM has $S$ scales. On scale $s\in{1,2,...,S}$, the feature map extracted by SP is split into $2^{s-1}$ strips on height dimension, i.e. $\sum_{s=1}^S2^{s-1}$ strips in total. Then a Global Pooling is applied to the 3-D strips to get 1-D features. For a strip $z_{s,t}$ where $t\in1,2,...,2^{s-1}$ stands index of the strip in the scale, the Global Pooling is formulated as 
$f'_{s,t}=maxpool(z_{s,t})+avgpool(z_{s,t})$, 
where $maxpool$ and $avgpool$ denote Global Max Pooling and Global Average Pooling respectively. Note that the functions $maxpool$ and $avgpool$ are used at the same time because it outperforms applying anyone of them alone. The final step is to employ FCs to map the features $f'$ into a discriminative space. Since strips in different scales depict features of different receptive fields, and different strips in each scales depict features of different spatial positions, it comes naturally to use independent FCs, as shown in Fig.~\ref{fig:hpm}.

\subsection{Multilayer Global Pipeline}\label{sec:MGP}

Different layers of a convolutional network have different receptive fields. The deeper the layer is, the larger the receptive field will be. Thus, pixels in feature maps of a shallow layer focus on local and fine-grained information while those in a deeper layer focus on more global and coarse-grained information. The set-level features extracted by applying SP on different layers have analogical property. As shown in the main pipeline of Fig.~\ref{fig:pipeline}, there is only one SP on the last layer of the convolutional network. To collect various-level \emph{set} information, Multilayer Global Pipeline~(MGP) is proposed. It has a similar structure with the convolutional network in the main pipeline and the set-level features extracted in different layers are added to MGP. The final feature map generated by MGP will also be mapped into $\sum_{s=1}^S2^{s-1}$ features by HPM. Note that the HPM after MGP does not share parameters with the HPM after the main pipeline.

\subsection{Training And Testing}\label{sec:TT}
\textbf{Training Loss}~~
As aforementioned, the output of the network is $2\times\sum_{s=1}^S2^{s-1}$ features with dimension $d$. The corresponding features among different samples will be used to compute the loss. In this paper, Batch All~($BA_+$) triplet loss is employed to train the network~\cite{hermans2017defense}. 
A batch with size of $p\times k$ is sampled from the training set where $p$ denotes the number of persons and $k$ denotes the number of training samples each person has in the batch.
Note that although the experiment shows that our model performs well when it is fed with the set composed by silhouettes gathered from arbitrary sequences, a sample used for training is actually composed by silhouettes sampled in one sequence. 

\textbf{Testing}~~
Given a query $\mathcal{Q}$, the goal is to retrieve all the \emph{sets} with the same identity in gallery set $\mathbb{G}$. Denote the sample in $\mathbb{G}$ as $\mathcal{G}$. The $\mathcal{Q}$ is first put into GaitSet net to generate multi-scale features, followed by concatenating all these features into a final representations $\mathcal{F}_\mathcal{Q}$ as shown in Fig.~\ref{fig:pipeline}. The same process is applied on each $\mathcal{G}$ to get $\mathcal{F}_\mathcal{G}$. Finally, $\mathcal{F}_\mathcal{Q}$ is compared with every $\mathcal{F}_\mathcal{G}$ using Euclidean distance to calculate Rank $1$ recognition accuracy.

\ifx\allinone\undefined
\newpage
\bibliographystyle{aaai}
\bibliography{../ref}
\end{document}
\fi
\ifx\allinone\undefined
\input{../preamble}
\begin{document}
\fi

\section{Experiments}
\label{sec:exp}
\begin{table*}[htbp]
\centering
\caption{Averaged rank-1 accuracies on \textbf{CASIA-B} under three different experimental settings, excluding identical-view cases.}
\fontsize{8}{9}\selectfont
\begin{tabularx}{\textwidth}{@{\extracolsep{\fill}}|c|c||c|p{0.4cm}|p{0.4cm}|p{0.4cm}|p{0.4cm}|p{0.4cm}|p{0.4cm}|p{0.4cm}|p{0.4cm}|p{0.4cm}|p{0.4cm}|p{0.4cm}|c|}
    \hline
    \multicolumn{3}{|c|}{Gallery NM\#1-4}&\multicolumn{11}{c|}{0\degree-180\degree}&\multirow{2}{*}{mean}\\ \cline{1-14}
    \multicolumn{1}{|c}{}&\multicolumn{1}{c}{Probe}&&0\degree&18\degree&36\degree&54\degree&72\degree&90\degree&108\degree&126\degree&144\degree&162\degree&180\degree&\\
    \hline \hline

    \multirow{8}{*}{\shortstack{ST \\ (24)}}&\multirow{6}{*}{NM\#5-6}&ViDP~\protect\cite{hu2013view} & $-$ & $-$ & $-$ & 59.1  & $-$ & 50.2  & $-$ & 57.5  & $-$ & $-$ & $-$ & $-$ \\
    &&CMCC~\protect\cite{kusakunniran2014recognizing} & 46.3  & $-$ & $-$ & 52.4  & $-$ & 48.3  & $-$ & 56.9  & $-$ & $-$ & $-$ & $-$ \\
    &&CNN-LB~\protect\cite{wu2017comprehensive}& 54.8  & $-$ & $-$ & 77.8  & $-$ & 64.9  & $-$ & 76.1  & $-$ & $-$ & $-$ & $-$ \\
    &&GaitSet(ours)&\textbf{64.6 } & \textbf{83.3 } & \textbf{90.4 } & \textbf{86.5 } & \textbf{80.2 } & \textbf{75.5 } & \textbf{80.3 } & \textbf{86.0 } & \textbf{87.1 } & \textbf{81.4 } & \textbf{59.6 } & \textbf{79.5 } \\
    \cline{2-15}
    &\multirow{1}{*}{BG\#1-2}&GaitSet(ours) & 55.8  & 70.5  & 76.9  & 75.5  & 69.7  & 63.4  & 68.0  & 75.8  & 76.2  & 70.7  & 52.5  & 68.6  \\
    \cline{2-15}
    &\multirow{1}{*}{CL\#1-2}&GaitSet(ours) & 29.4  & 43.1  & 49.5  & 48.7  & 42.3  & 40.3  & 44.9  & 47.4  & 43.0  & 35.7  & 25.6  & 40.9  \\
    \hline \hline

    \multirow{12}{*}{\shortstack{MT \\ (62)}}&\multirow{4}{*}{NM\#5-6}
    &AE~\protect\cite{yu2017invariant} & 49.3  & 61.5  & 64.4  & 63.6  & 63.7  & 58.1  & 59.9  & 66.5  & 64.8  & 56.9  & 44.0  & 59.3  \\
    &&MGAN~\protect\cite{he2019multi} & 54.9  & 65.9  & 72.1  & 74.8  & 71.1  & 65.7  & 70.0  & 75.6  & 76.2  & 68.6  & 53.8  & 68.1  \\
    &&GaitSet(ours) & \textbf{86.8 } & \textbf{95.2 } & \textbf{98.0 } & \textbf{94.5 } & \textbf{91.5 } & \textbf{89.1 } & \textbf{91.1 } & \textbf{95.0 } & \textbf{97.4 } & \textbf{93.7 } & \textbf{80.2 } & \textbf{92.0 } \\
    \cline{2-15}
    &\multirow{4}{*}{BG\#1-2}
    &AE~\protect\cite{yu2017invariant} & 29.8  & 37.7  & 39.2  & 40.5  & 43.8  & 37.5  & 43.0  & 42.7  & 36.3  & 30.6  & 28.5  & 37.2  \\
    &&MGAN~\protect\cite{he2019multi} & 48.5  & 58.5  & 59.7  & 58.0  & 53.7  & 49.8  & 54.0  & 61.3  & 59.5  & 55.9  & 43.1  & 54.7  \\
    &&GaitSet(ours) & \textbf{79.9 } & \textbf{89.8 } & \textbf{91.2 } & \textbf{86.7 } & \textbf{81.6 } & \textbf{76.7 } & \textbf{81.0 } & \textbf{88.2 } & \textbf{90.3 } & \textbf{88.5 } & \textbf{73.0 } & \textbf{84.3 } \\
    \cline{2-15}
    &\multirow{4}{*}{CL\#1-2}
    &AE~\protect\cite{yu2017invariant} & 18.7  & 21.0  & 25.0  & 25.1  & 25.0  & 26.3  & 28.7  & 30.0  & 23.6  & 23.4  & 19.0  & 24.2  \\
    &&MGAN~\protect\cite{he2019multi} & 23.1  & 34.5  & 36.3  & 33.3  & 32.9  & 32.7  & 34.2  & 37.6  & 33.7  & 26.7  & 21.0  & 31.5  \\
    &&GaitSet(ours) & \textbf{52.0 } & \textbf{66.0 } & \textbf{72.8 } & \textbf{69.3 } & \textbf{63.1 } & \textbf{61.2 } & \textbf{63.5 } & \textbf{66.5 } & \textbf{67.5 } & \textbf{60.0 } & \textbf{45.9 } & \textbf{62.5 } \\
    \hline \hline

    \multirow{7}{*}{\shortstack{LT \\ (74)}}&\multirow{3}{*}{NM\#5-6}&CNN-3D~\protect\cite{wu2017comprehensive}&87.1  & 93.2  & 97.0  & 94.6  & 90.2  & 88.3  & 91.1  & 93.8  & 96.5  & 96.0  & 85.7  & 92.1  \\
    &&CNN-Ensemble~\protect\cite{wu2017comprehensive} &88.7  & 95.1  & 98.2  & 96.4  & \textbf{94.1 } & 91.5  & 93.9  & 97.5  & 98.4  & 95.8  & 85.6  & 94.1  \\
    &&GaitSet(ours) & \textbf{90.8 } & \textbf{97.9 } & \textbf{99.4 } & \textbf{96.9 } & 93.6  & \textbf{91.7 } & \textbf{95.0 } & \textbf{97.8 } & \textbf{98.9 } & \textbf{96.8 } & \textbf{85.8 } & \textbf{95.0 } \\
    \cline{2-15}
    &\multirow{2}{*}{BG\#1-2}&CNN-LB~\protect\cite{wu2017comprehensive}&64.2  & 80.6  & 82.7  & 76.9  & 64.8  & 63.1  & 68.0  & 76.9  & 82.2  & 75.4  & 61.3  & 72.4  \\
    &&GaitSet(ours) & \textbf{83.8 } & \textbf{91.2 } & \textbf{91.8 } & \textbf{88.8 } & \textbf{83.3 } & \textbf{81.0 } & \textbf{84.1 } & \textbf{90.0 } & \textbf{92.2 } & \textbf{94.4 } & \textbf{79.0 } & \textbf{87.2 } \\
    \cline{2-15}
    &\multirow{2}{*}{CL\#1-2}&CNN-LB~\protect\cite{wu2017comprehensive}&37.7  & 57.2  & 66.6  & 61.1  & 55.2  & 54.6  & 55.2  & 59.1  & 58.9  & 48.8  & 39.4  & 54.0  \\
    &&GaitSet(ours) & \textbf{61.4 } & \textbf{75.4 } & \textbf{80.7 } & \textbf{77.3 } & \textbf{72.1 } & \textbf{70.1 } & \textbf{71.5 } & \textbf{73.5 } & \textbf{73.5 } & \textbf{68.4 } & \textbf{50.0 } & \textbf{70.4 } \\
    \hline

\end{tabularx}
\label{tab:casia-b}
\end{table*}
Our empirical experiments mainly contain three parts. The first part compares GaitSet with other state-of-the-art methods on two public gait datasets: CASIA-B~\cite{yu2006framework} and OU-MVLP~\cite{Takemura2018}. The Second part is ablation experiments conducted on CASIA-B. 
In the third part, we investigated the practicality of GaitSet in three aspects: the performance on limited silhouettes, multiple views and multiple walking conditions.
\subsection{Datasets and Training Details}
\textbf{CASIA-B}~dataset~\cite{yu2006framework} is a popular gait dataset. It contains 124 subjects~(labeled in 001-124), 3 walking conditions and 11 views ($0^\circ, 18^\circ, ..., 180^\circ$). The walking condition contains normal~(NM) (6 sequences per subject), walking with bag~(BG) (2 sequences per subject) and wearing coat or jacket~(CL) (2 sequences per subject). Namely, each subject has $11\times(6+2+2)=110$ sequences. 
As there is no official partition of training and test sets of this dataset, we conduct experiments on three settings which are popular in current literatures. We name these three settings as small-sample training~(ST), medium-sample training~(MT) and large-sample training~(LT). In ST, the first 24 subjects~(labeled in 001-024) are used for training and the rest 100 subjects are leaved for test. In MT, the first 62 subjects are used for training and the rest 62 subjects are leaved for test. In LT, the first 74 subjects are used for training and the rest 50 subjects are leaved for test. In the test sets of all three settings, the first 4 sequences of the NM condition~(NM \#1-4) are kept in gallery, and the rest 6 sequences are divided into 3 probe subsets, i.e. NM subsets containing NM \#5-6, BG subsets containing BG \#1-2 and CL subsets containing CL \#1-2.

\textbf{OU-MVLP}~dataset~\cite{Takemura2018} is so far the world's largest public gait dataset. It contains 10,307 subjects, 14 views~($0^\circ, 15^\circ, ...,90^\circ;180^\circ,195^\circ,...,270^\circ$) per subject and 2 sequences~(\#00-01) per view. The sequences are divided into training and test set by subjects~(5153 subjects for training and 5154 subjects for test).
In the test set, sequences with index \#01 are kept in gallery and those with index \#00 are used as probes.

\textbf{Training Details}~~In all the experiments, the input is a set of aligned silhouettes in size of $64\times44$. The silhouettes are directly provided by the datasets and are aligned based on methods in~\cite{Takemura2018}. The set cardinality in the training is set to be $30$. Adam is chosen as an optimizer~\cite{kingma2014adam}. 
The number of scales $S$ in HPM is set as $5$. The margin in $BA_+$ triplet loss is set as $0.2$. 
The models are trained with 8 NVIDIA 1080TI GPUs. \textbf{1)}~In CASIA-B, the mini-batch is composed by the manner introduced in Sec.~\ref{sec:TT} with $p=8$ and $k=16$. We set the number of channels in $C1$ and $C2$ as 32, in $C3$ and $C4$ as 64 and in $C5$ and $C6$ as 128. Under this setting, the average computational complexity of our model is 8.6GFLOPs. The learning rate is set to be $1e-4$. For ST, we train our model for 50K iterations. For MT, we train it for 60K iterations. For LT, we train it for 80K iterations. \textbf{2)}~In OU-MVLP, since it contains 20 times more sequences than CASIA-B, we use convolutional layers with more channels~($C1=C2=64, C3=C4=128, C5=C6=256$) and train it with larger batch size~($p=32,k=16$). 
The learning rate is $1e-4$ in the first 150K iterations, and then is changed into $1e-5$ for the rest of 100K iterations.

\subsection{Main Results}
\label{sec:mr}
\subsubsection{CASIA-B}
Tab.~\ref{tab:casia-b} shows the comparison between the state-of-the-art methods~\footnote{Since~\cite{wu2017comprehensive} proposed more than one model, the most competitive results under different experimental settings are cited.} and our GaitSet. Except of ours, other results are directly taken from their original papers. 
All the results are averaged on the 11 gallery views and the identical views are excluded. For example, the accuracy of probe view $36^\circ$ is averaged on 10 gallery views, excluding gallery view $36^\circ$. An interesting pattern between views and accuracies can be observed in Tab.~\ref{tab:casia-b}. Besides $0^\circ$ and $180^\circ$
, the accuracy of $90^\circ$ is a local minimum value. It is always worse than that of $72^\circ$ or $108^\circ$. The possible reason is that gait information contains not only those parallel to the walking direction like stride which can be observed most clearly at $90^\circ$, but also those vertical to the walking direction like a left-right swinging of body or arms which can be observed most clearly at $0^\circ$ or $180^\circ$. So, both parallel and vertical perspectives lose some part of gait information while views like $36^\circ$ or $144^\circ$ can obtain most of it.

\textbf{Small-Sample Training~(ST)}~~Our method achieves a high performance even with only 24 subjects in the training set and exceed the best performance reported so far~\cite{wu2017comprehensive} over 10 percent on the views they reported. There are mainly two reasons. \textbf{1)} As our model regards the input as a set, images used to train the convolution network in the main pipeline are dozens of times more than those models based on gait templates. Taking a mini-batch for an example, our model is fed with $30\times128=3840$ silhouettes while under the same batch size models using gait templates can only get $128$ templates. \textbf{2)} Since the sample \emph{sets} used in training phase are composed by frames selected randomly from the sequence, each sequence in the training set can generate multiple different \emph{sets}. Thus any units related to set feature learning like MGP and HPM can also be trained well. 


\begin{table*}[t]
\centering
\caption{Ablation experiments conducted on \textbf{CASIA-B} using setting LT. Results are rank-1 accuracies averaged on all 11 views, excluding identical-view cases. The numbers in brackets indicate the second highest results in each column.}
\fontsize{8}{9}\selectfont
    \begin{tabularx}{\textwidth}{@{\extracolsep{\fill}}|c|c|cccccc|cc|c||c|c|c|}
    \hline
    \multirow{2}{*}{GEI} & \multirow{2}{*}{Set} & \multicolumn{6}{c|}{Set Pooling}               & \multicolumn{2}{c|}{HPM weight} & \multirow{2}{*}{MGP} & \multirow{2}{*}{NM} & \multirow{2}{*}{BG} & \multirow{2}{*}{CL} \\
    \cline{3-10}
          &       & Max   & Mean  & Median & Joint sum~\ref{eq:sum} & Joint 1\_1C~\ref{eq:cat} & Attention & Shared & Independent &       &       &       &  \\
    \hline
    $\surd$ &       &       &       &       &       &       &       & $\surd$ &       &       & 80.4  & 68.1  & 40.8  \\
    \hline
          & $\surd$ & $\surd$ &       &       &       &       &       & $\surd$ &       &       & 91.3  & 82.3  & 67.1  \\
    \hline
          & $\surd$ & $\surd$ &       &       &       &       &       &       & $\surd$ &       & 93.2  & 84.7  & (70.2)\\
    \hline
          & $\surd$ &       & $\surd$ &       &       &       &       &       & $\surd$ &       & 90.0  & 79.5  & 57.1  \\
    \hline
          & $\surd$ &       &       & $\surd$ &       &       &       &       & $\surd$ &       & 89.5  & 78.1  & 53.5  \\
    \hline
          & $\surd$ &       &       &       & $\surd$ &       &       &       & $\surd$ &       & 92.4  & 82.8  & 63.4  \\
    \hline
          & $\surd$ &       &       &       &       & $\surd$ &       &       & $\surd$ &       & 93.3  & (85.7)& 66.3  \\
    \hline
          & $\surd$ &       &       &       &       &       & $\surd$ &       & $\surd$ &       & (93.7)& 84.2  & 69.4  \\
    \hline
          & \textbf{$\surd$} & \textbf{$\surd$} &       &       &       &       &       &       & $\surd$ & \textbf{$\surd$} & \textbf{95.0 } & \textbf{87.2 } & \textbf{70.4 } \\
    \hline
    \end{tabularx}
  \label{tab:ablation}
\end{table*}

\begin{table}[htb]
\centering
\caption{Averaged rank-1 accuracies on \textbf{OU-MVLP}, excluding identical-view cases. GEINet:~\protect\cite{shiraga2016geinet}. 3in+2diff:~\protect\cite{takemura2017input}}.
\fontsize{8}{9}\selectfont
\begin{tabularx}{\columnwidth}{|c|c|@{\extracolsep{\fill}}c|c|c|c|}
    \hline
    \multirow{2}{*}{Probe} & \multicolumn{2}{c|}{Gallery All 14 Views} & \multicolumn{3}{c|}{Gallery $0^\circ, 30^\circ, 60^\circ, 90^\circ$} \\
    \cline{2-6}
          & GEINet & Ours & GEINet & 3in+2diff & Ours \\
    \hline
    $0^\circ$     & 11.4  & \textbf{79.5 } & 8.2   & 25.5  & \textbf{77.7 } \\
    \hline
    $15^\circ$    & 29.1  & \textbf{87.9 } & -     & -     & \textbf{86.3 } \\
    \hline
    $30^\circ$    & 41.5  & \textbf{89.9 } & 32.3  & 50.0  & \textbf{86.9 } \\
    \hline
    $45^\circ$    & 45.5  & \textbf{90.2 } & -     & -     & \textbf{89.1 } \\
    \hline
    $60^\circ$    & 39.5  & \textbf{88.1 } & 33.6  & 45.3  & \textbf{85.3 } \\
    \hline
    $75^\circ$    & 41.8  & \textbf{88.7 } & -     & -     & \textbf{87.6 } \\
    \hline
    $90^\circ$    & 38.9  & \textbf{87.8 } & 28.5  & 40.6  & \textbf{83.5 } \\
    \hline
    $180^\circ$   & 14.9  & \textbf{81.7 } & -     & -     & \textbf{80.5 } \\
    \hline
    $195^\circ$   & 33.1  & \textbf{86.7 } & -     & -     & \textbf{82.8 } \\
    \hline
    $210^\circ$   & 43.2  & \textbf{89.0 } & -     & -     & \textbf{87.2 } \\
    \hline
    $225^\circ$   & 45.6  & \textbf{89.3 } & -     & -     & \textbf{86.8 } \\
    \hline
    $240^\circ$   & 39.4  & \textbf{87.2 } & -     & -     & \textbf{85.4 } \\
    \hline
    $255^\circ$   & 40.5  & \textbf{87.8 } & -     & -     & \textbf{85.7 } \\
    \hline
    $270^\circ$   & 36.3  & \textbf{86.2 } & -     & -     & \textbf{85.0 } \\
    \hline
    mean  & 35.8  & \textbf{87.1 } & -     & -     & \textbf{85.0 } \\
    \hline

\end{tabularx}
\label{tab:oumvlp}
\end{table}

\textbf{Medium-Sample Training~(MT) \& Large-Sample Training~(LT)}~~
Tab.~\ref{tab:casia-b} shows that our model obtains very nice results on the NM subset, especially on LT where results of all views except $180^\circ$ are over $90\%$. On the BG and CL subsets, although the accuracies of some views like $0^\circ$ and $180^\circ$ are still not high, the mean accuracies of our model exceed those of other models for at least $18.8\%$.

\subsubsection{OU-MVLP}

Tab.~\ref{tab:oumvlp} shows our results. As some of the previous works did not conduct experiments on all 14 views, we list our results on two kinds of gallery sets, i.e. all 14 views and 4 typical views ($0^\circ, 30^\circ， 60^\circ， 90^\circ$). All the results are averaged on the gallery views and the identical views are excluded. The results show that our methods can generalize well on the dataset with such a large scale and wide view variation. Further, since representation for each sample only needs to be calculated once, our model can complete the test~(containing 133780 sequences) in only 7 minutes with 8 NVIDIA 1080TI GPUs. It is note worthy that since some subjects miss several gait sequences and we did not remove them from the probe, the maximum of rank-1 accuracy cannot reach $100\%$. If we ignore the cases which have no corresponding samples in the gallery, the average rank-1 accuracy of all probe views is $93.3\%$ rather than $87.1\%$.

\subsection{Ablation Experiments}
\label{sec:abl}
Tab.~\ref{tab:ablation} shows the thorough results of ablation experiments. The effectiveness of every innovation in Sec.~\ref{sec:method} is studied.

\textbf{Set VS. GEI}~~
The first two lines of Tab.~\ref{tab:ablation} show the effectiveness of regarding gait as a set. With fully identical networks, the result of using set exceeds that of using GEI by more than $10\%$ on NM subset and more than $25\%$ on CL subset. The only difference is that in GEI experiment, gait silhouettes are averaged into a single GEI before being fed into the network. 
There are mainly two reasons for this phenomenal improvement. \textbf{1)}~
Our SP extracts the set-level feature based on high-level feature map where temporal information can be well preserved and spatial information has been sufficiently processed. \textbf{2)}~As mentioned in Sec.~\ref{sec:mr}, regarding gait as a set enlarges the volume of training data.

\textbf{Impact of SP}~~
In Tab.~\ref{tab:ablation}, the results from the third line to the eighth line show the impact of different SP strategies. SP with attention, $1\times1$ convolution~(1\_1C) joint function and $\mathrm{max}(\cdot)$ obtain the highest accuracy on the NM, BG, and CL subsets respectively. Considering SP with $\mathrm{max}(\cdot)$ also achieved the second best performance on the NM and BG subset and has the most concise structure, we choose it as SP in the final version of GaitSet.

\textbf{Impact of HPM and MGP}~~
The second and the third lines of Tab.~\ref{tab:ablation} compare the impact of independent weight in HPM. It can be seen that using independent weight improves the accuracy by about $2\%$ on each subset. In the experiments, we also find out that the introduction of independent weight helps the network converge faster.
The last two lines of Tab.~\ref{tab:ablation} show that MGP can bring improvement on all three test subsets. This result is consistent the theory mentioned in Sec.~\ref{sec:MGP} that set-level features extracted from different layers of the main pipeline contain different valuable information.

\subsection{Practicality}
\label{sec:pra}
Due to the flexibility of set, GaitSet has great potential in more complicated practical conditions. In this section, we investigate the practicality of GaitSet through three novel scenarios. \textbf{1)}~How will it perform when the input set only contains a few silhouettes? \textbf{2)}~Can silhouettes with different views enhance the identification accuracy? \textbf{3)}~Whether can the model effectively extract discriminative representation from a set containing silhouettes shot under different walking conditions. It is worth noting that we did not retrain our model in these experiments. It is fully identical to that in Sec.~\ref{sec:mr} with setting LT. Note that, all the experiments containing random selection in this section are ran for $10$ times and the average accuracies are reported.

\begin{figure}[htb]
\centering
\includegraphics[width=1\linewidth, clip=true, trim=230 140 230 140]{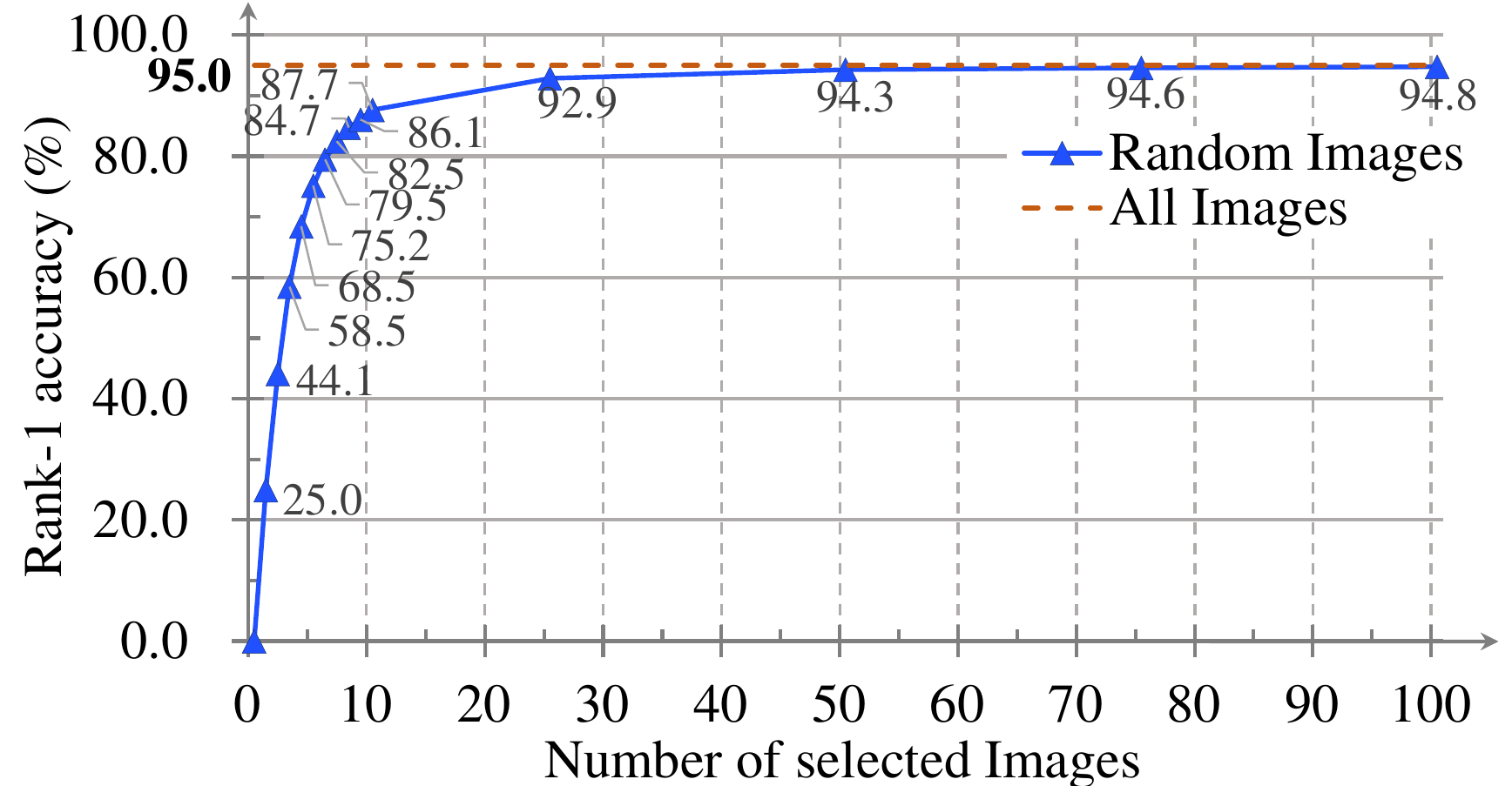}
\caption{Average rank-1 accuracies with constraints of silhouette volume on \textbf{CASIA-B} using setting LT. Accuracies are averaged on all 11 views excluding identical-view cases, and the final reported results are averaged across 10 times experiments.}
\label{fig:frames}
\end{figure}
\textbf{Limited Silhouettes}~~In real forensic identification scenarios, there are cases that we do not have a continuous sequence of a subject's gait but only some fitful and sporadic silhouettes. We simulate such a circumstance by randomly selecting a certain number of frames from sequences to compose each sample in both gallery and probe. Fig.~\ref{fig:frames} shows the relationship between the number of silhouettes in each input set and the rank-1 accuracy averaged on all 11 probe views. Our method attains an $82\%$ accuracy with only 7 silhouettes. The result also indicates that our model makes full use of the temporal information of gait. Since \textbf{1)} the accuracy rises monotonically with the increase of the number of silhouettes. \textbf{2)} The accuracy is close to the best performance when the samples contain more than 25 silhouettes. This number is consistent with the number of frames that one gait period contains.

\textbf{Multiple Views}~~
There are conditions that different views of one person's gait can be gathered.
We simulate these scenarios by constructing each sample with silhouettes selected from two sequences with the same walking condition but different views. To eliminate the effects of silhouette number, we also conduct an experiment in which the silhouette number is limited to 10. Specifically, in the contrast experiments of single view, an input set is composed by 10 silhouettes from one sequence. In the two-view experiment, an input set is composed by 5 silhouettes from each of two sequences. Note that in this experiment, only probe samples are composed by the way discussed above, whereas sample in the gallery is composed by all silhouettes from one sequence.

\begin{table}[htbp]
\centering
\caption{Multi-view experiments conducted on \textbf{CASIA-B} using setting LT. Cases where the probe contains the view of the gallery are excluded.}
\fontsize{8}{9}\selectfont
    \begin{tabular}{|c|c|c|c|c|c||c|}
    \hline
    \multirow{2}{*}{\shortstack{View \\ difference}} & 
    \multirow{2}{*}{\shortstack{$18^\circ$/\\$162^\circ$}} &
    \multirow{2}{*}{\shortstack{$36^\circ$/\\$144^\circ$}} &
    \multirow{2}{*}{\shortstack{$54^\circ$/\\$126^\circ$}} &
    \multirow{2}{*}{\shortstack{$72^\circ$/\\$108^\circ$}} &
    \multirow{2}{*}{$90^\circ$} & 
    \multirow{2}{*}{\shortstack{Single \\ view}} \\
    &&&&&&\\
    \hline
    All silhouettes & 97.0  & 97.9  & 98.7  & 99.1  & 99.0  & 95.0  \\
    \hline
    10 silhouettes  & 87.9  & 90.6  & 92.7  & 93.7  & 93.7  & 87.7  \\
    \hline
    \end{tabular}%
  \label{tab:mv}%
\end{table}%

Tab.~\ref{tab:mv} shows the results. As there are too many view pairs to be shown, we summarize the results by averaging accuracies of each possible view difference. For example, the result of $90^\circ$ difference is averaged by accuracies of $6$ view pairs~($0^\circ\&90^\circ, 18^\circ\&108^\circ, ..., 90^\circ\&180^\circ$). Further, the 9 view differences are folded at $90^\circ$ and those larger than $90^\circ$ are averaged with the corresponding view differences less than $90^\circ$. For example, the results of $18^\circ$ view difference are averaged with those of $162^\circ$ view difference. It can be seen that our model can aggregate information from different views and boost the performance.
This can be explained by the pattern between views and accuracies that we have discussed in Sec.~\ref{sec:mr}. Containing multiple views in the input set can let the model gather both parallel and vertical information, resulting in performance improvement.

\textbf{Multiple Walking Conditions}~~
In real life, it is highly possible that gait sequences of the same person
 are under different walking conditions.
 We simulate such a condition by forming input set with silhouettes from two sequences with same view but different walking conditions. We conduct experiments with different silhouette number constraints. Note that in this experiment, only probe samples are composed by the way discussed above. Any sample in the gallery is constituted by all silhouettes from one sequence. What's more, the probe-gallery division of this experiment is different.
 For each subject, sequences NM \#02, BG \#02 and CL \#02 are kept in the gallery and sequences NM \#01, BG \#01 and CL \#01 are used as probe.

\begin{table}[htbp]
\centering
\caption{Multiple walking condition experiments conducted on \textbf{CASIA-B} using setting LT. Results are rank-1 accuracies averaged on all 11 views, excluding identical-view cases. The numbers in brackets indicate the constraints of silhouette number in each input set.}
\fontsize{8}{9}\selectfont
    \begin{tabularx}{\columnwidth}{@{\extracolsep{\fill}}|c|c|c|c|c|c|}
    \hline
    NM(10) & 81.5  & NM(10)+BG(10) & 87.9  & NM(20) & 89.8  \\
    \hline
    BG(10) & 77.1  & NM(10)+CL(10) & 85.8  & BG(20) & 84.1  \\
    \hline
    CL(10) & 74.4  & BG(10)+CL(10) & 84.6  & CL(20) & 82.6  \\
    \hline
    \end{tabularx}%
  \label{tab:mw}%
\end{table}%

Tab.~\ref{tab:mw} shows the results. First, the accuracies will still be boosted with the increase of silhouette number. Second, when the number of silhouettes are fixed, the results reveal relationships between different walking conditions. Silhouettes of BG and CL contain massive but different noises, which makes them complementary with each other. Thus, their combination can improve the accuracy. However, silhouettes of NM contain few noises, so substituting some of them with silhouettes of other two conditions cannot bring extra information but only noises and can decrease the accuracies. 

\ifx\allinone\undefined
\newpage
\bibliographystyle{aaai}
\bibliography{../ref}
\end{document}
\fi
\ifx\allinone\undefined
\input{../preamble}
\begin{document}
\fi

\section{Conclusion}
\label{sec:con}
In this paper, we presented a novel perspective that regards gait as a set and thus proposed a GaitSet approach. The GaitSet can extract both spatial and temporal information more effectively and efficiently than those existing methods regarding gait as a template or sequence.
It also provide a novel way to aggregate valuable information from different sequences to enhance the recognition accuracy.
Experiments on two benchmark gait datasets has indicated that compared with other state-of-the-art algorithms, GaitSet achieves the highest recognition accuracy, and reveals a wide range of flexibility on various complex environments, showing a great potential in practical applications. In the future, we will investigate a more effective instantiation for Set Pooling~(SP) and further improve the performance in complex scenarios.

\ifx\allinone\undefined
\end{document}
\fi

\bibliographystyle{aaai}
\bibliography{ref}

\begin{thebibliography}{}

\bibitem[\protect\citeauthoryear{Bashir, Xiang, and
  Gong}{2010}]{bashir2010gait}
Bashir, K.; Xiang, T.; and Gong, S.
\newblock 2010.
\newblock Gait recognition without subject cooperation.
\newblock {\em Pattern Recognition Letters} 31(13):2052--2060.

\bibitem[\protect\citeauthoryear{Charles \bgroup et al\mbox.\egroup
  }{2017}]{qi2017pointnet}
Charles, R.~Q.; Su, H.; Kaichun, M.; and Guibas, L.~J.
\newblock 2017.
\newblock Pointnet: Deep learning on point sets for 3{D} classification and
  segmentation.
\newblock In {\em CVPR},  77--85.

\bibitem[\protect\citeauthoryear{Fu \bgroup et al\mbox.\egroup
  }{2018}]{fu2018horizontal}
Fu, Y.; Wei, Y.; Zhou, Y.; Shi, H.; Huang, G.; Wang, X.; Yao, Z.; and Huang, T.
\newblock 2018.
\newblock Horizontal pyramid matching for person re-identification.
\newblock {\em ArXiv:1804.05275}.

\bibitem[\protect\citeauthoryear{Hamilton, Ying, and
  Leskovec}{2017}]{hamilton2017inductive}
Hamilton, W.; Ying, Z.; and Leskovec, J.
\newblock 2017.
\newblock Inductive representation learning on large graphs.
\newblock In {\em NIPS},  1024--1034.

\bibitem[\protect\citeauthoryear{Han and Bhanu}{2006}]{han2006individual}
Han, J., and Bhanu, B.
\newblock 2006.
\newblock Individual recognition using gait energy image.
\newblock {\em IEEE TPAMI} 28(2):316--322.

\bibitem[\protect\citeauthoryear{He \bgroup et al\mbox.\egroup
  }{2019}]{he2019multi}
He, Y.; Zhang, J.; Shan, H.; and Wang, L.
\newblock 2019.
\newblock Multi-task {GAN}s for view-specific feature learning in gait
  recognition.
\newblock {\em IEEE TIFS} 14(1):102--113.

\bibitem[\protect\citeauthoryear{Hermans, Beyer, and
  Leibe}{2017}]{hermans2017defense}
Hermans, A.; Beyer, L.; and Leibe, B.
\newblock 2017.
\newblock In defense of the triplet loss for person re-identification.
\newblock {\em ArXiv:1703.07737}.

\bibitem[\protect\citeauthoryear{Hu \bgroup et al\mbox.\egroup
  }{2013}]{hu2013view}
Hu, M.; Wang, Y.; Zhang, Z.; Little, J.~J.; and Huang, D.
\newblock 2013.
\newblock View-invariant discriminative projection for multi-view gait-based
  human identification.
\newblock {\em IEEE TIFS} 8(12):2034--2045.

\bibitem[\protect\citeauthoryear{Kingma and Ba}{2015}]{kingma2014adam}
Kingma, D.~P., and Ba, J.
\newblock 2015.
\newblock Adam: A method for stochastic optimization.
\newblock {\em ICLR}.

\bibitem[\protect\citeauthoryear{Krause \bgroup et al\mbox.\egroup
  }{2017}]{krause2017hierarchical}
Krause, J.; Johnson, J.; Krishna, R.; and Fei-Fei, L.
\newblock 2017.
\newblock A hierarchical approach for generating descriptive image paragraphs.
\newblock In {\em CVPR},  3337--3345.

\bibitem[\protect\citeauthoryear{Kusakunniran \bgroup et al\mbox.\egroup
  }{2014}]{kusakunniran2014recognizing}
Kusakunniran, W.; Wu, Q.; Zhang, J.; Li, H.; and Wang, L.
\newblock 2014.
\newblock Recognizing gaits across views through correlated motion
  co-clustering.
\newblock {\em IEEE TIP} 23(2):696--709.

\bibitem[\protect\citeauthoryear{Li, Zhu, and Gong}{2018}]{li2018harmonious}
Li, W.; Zhu, X.; and Gong, S.
\newblock 2018.
\newblock Harmonious attention network for person re-identification.
\newblock In {\em CVPR}.

\bibitem[\protect\citeauthoryear{Liao \bgroup et al\mbox.\egroup
  }{2017}]{liao2017pose}
Liao, R.; Cao, C.; Garcia, E.~B.; Yu, S.; and Huang, Y.
\newblock 2017.
\newblock Pose-based temporal-spatial network (ptsn) for gait recognition with
  carrying and clothing variations.
\newblock In {\em Chinese Conference on Biometric Recognition},  474--483.
\newblock Springer.

\bibitem[\protect\citeauthoryear{Makihara \bgroup et al\mbox.\egroup
  }{2006}]{makihara2006gait}
Makihara, Y.; Sagawa, R.; Mukaigawa, Y.; Echigo, T.; and Yagi, Y.
\newblock 2006.
\newblock Gait recognition using a view transformation model in the frequency
  domain.
\newblock In {\em ECCV},  151--163.
\newblock Springer.

\bibitem[\protect\citeauthoryear{Qi \bgroup et al\mbox.\egroup
  }{2017}]{qi2017pointnet++}
Qi, C.~R.; Yi, L.; Su, H.; and Guibas, L.~J.
\newblock 2017.
\newblock {PointNet++}: Deep hierarchical feature learning on point sets in a
  metric space.
\newblock In {\em NIPS},  5099--5108.

\bibitem[\protect\citeauthoryear{Shiraga \bgroup et al\mbox.\egroup
  }{2016}]{shiraga2016geinet}
Shiraga, K.; Makihara, Y.; Muramatsu, D.; Echigo, T.; and Yagi, Y.
\newblock 2016.
\newblock {GEINet}: View-invariant gait recognition using a convolutional
  neural network.
\newblock In {\em ICB},  1--8.

\bibitem[\protect\citeauthoryear{Takemura \bgroup et al\mbox.\egroup
  }{2018a}]{takemura2017input}
Takemura, N.; Makihara, Y.; Muramatsu, D.; Echigo, T.; and Yagi, Y.
\newblock 2018a.
\newblock On input/output architectures for convolutional neural network-based
  cross-view gait recognition.
\newblock {\em IEEE TCSVT} 28(1):1--13.

\bibitem[\protect\citeauthoryear{Takemura \bgroup et al\mbox.\egroup
  }{2018b}]{Takemura2018}
Takemura, N.; Makihara, Y.; Muramatsu, D.; Echigo, T.; and Yagi, Y.
\newblock 2018b.
\newblock Multi-view large population gait dataset and its performance
  evaluation for cross-view gait recognition.
\newblock {\em IPSJ TCVA} 10(4):1--14.

\bibitem[\protect\citeauthoryear{Wang \bgroup et al\mbox.\egroup
  }{2012}]{wang2012human}
Wang, C.; Zhang, J.; Wang, L.; Pu, J.; and Yuan, X.
\newblock 2012.
\newblock Human identification using temporal information preserving gait
  template.
\newblock {\em IEEE TPAMI} 34(11):2164--2176.

\bibitem[\protect\citeauthoryear{Wang \bgroup et al\mbox.\egroup
  }{2018a}]{wang2018learning}
Wang, G.; Yuan, Y.; Chen, X.; Li, J.; and Zhou, X.
\newblock 2018a.
\newblock Learning discriminative features with multiple granularities for
  person re-identification.
\newblock {\em ArXiv:1804.01438}.

\bibitem[\protect\citeauthoryear{Wang \bgroup et al\mbox.\egroup
  }{2018b}]{wang2018non}
Wang, X.; Girshick, R.; Gupta, A.; and He, K.
\newblock 2018b.
\newblock Non-local neural networks.
\newblock In {\em CVPR}.

\bibitem[\protect\citeauthoryear{Wang \bgroup et al\mbox.\egroup
  }{2018c}]{wang2018dynamic}
Wang, Y.; Sun, Y.; Liu, Z.; Sarma, S.~E.; Bronstein, M.~M.; and Solomon, J.~M.
\newblock 2018c.
\newblock Dynamic graph {CNN} for learning on point clouds.
\newblock {\em ArXiv:1801.07829}.

\bibitem[\protect\citeauthoryear{Wolf, Babaee, and
  Rigoll}{2016}]{takemura2018multi}
Wolf, T.; Babaee, M.; and Rigoll, G.
\newblock 2016.
\newblock Multi-view gait recognition using 3{D} convolutional neural networks.
\newblock In {\em ICIP},  4165--4169.

\bibitem[\protect\citeauthoryear{Wu \bgroup et al\mbox.\egroup
  }{2017}]{wu2017comprehensive}
Wu, Z.; Huang, Y.; Wang, L.; Wang, X.; and Tan, T.
\newblock 2017.
\newblock A comprehensive study on cross-view gait based human identification
  with deep {CNN}s.
\newblock {\em IEEE TPAMI} 39(2):209--226.

\bibitem[\protect\citeauthoryear{Xing \bgroup et al\mbox.\egroup
  }{2016}]{xing2016complete}
Xing, X.; Wang, K.; Yan, T.; and Lv, Z.
\newblock 2016.
\newblock Complete canonical correlation analysis with application to
  multi-view gait recognition.
\newblock {\em Pattern Recognition} 50:107--117.

\bibitem[\protect\citeauthoryear{Xu \bgroup et al\mbox.\egroup
  }{2015}]{xu2015show}
Xu, K.; Ba, J.; Kiros, R.; Cho, K.; Courville, A.; Salakhudinov, R.; Zemel, R.;
  and Bengio, Y.
\newblock 2015.
\newblock Show, attend and tell: Neural image caption generation with visual
  attention.
\newblock In {\em ICML},  2048--2057.

\bibitem[\protect\citeauthoryear{Yu \bgroup et al\mbox.\egroup
  }{2017a}]{yu2017gaitgan}
Yu, S.; Chen, H.; Reyes, E. B.~G.; and Poh, N.
\newblock 2017a.
\newblock Gait{GAN}: Invariant gait feature extraction using generative
  adversarial networks.
\newblock In {\em CVPR Workshops},  532--539.

\bibitem[\protect\citeauthoryear{Yu \bgroup et al\mbox.\egroup
  }{2017b}]{yu2017invariant}
Yu, S.; Chen, H.; Wang, Q.; Shen, L.; and Huang, Y.
\newblock 2017b.
\newblock Invariant feature extraction for gait recognition using only one
  uniform model.
\newblock {\em Neurocomputing} 239:81--93.

\bibitem[\protect\citeauthoryear{Yu, Tan, and Tan}{2006}]{yu2006framework}
Yu, S.; Tan, D.; and Tan, T.
\newblock 2006.
\newblock A framework for evaluating the effect of view angle, clothing and
  carrying condition on gait recognition.
\newblock In {\em ICPR}, volume~4,  441--444.

\bibitem[\protect\citeauthoryear{Zaheer \bgroup et al\mbox.\egroup
  }{2017}]{zaheer2017deep}
Zaheer, M.; Kottur, S.; Ravanbakhsh, S.; Poczos, B.; Salakhutdinov, R.~R.; and
  Smola, A.~J.
\newblock 2017.
\newblock Deep sets.
\newblock In {\em NIPS},  3391--3401.

\bibitem[\protect\citeauthoryear{Zhou and Tuzel}{2018}]{zhou2017voxelnet}
Zhou, Y., and Tuzel, O.
\newblock 2018.
\newblock Voxelnet: End-to-end learning for point cloud based 3{D} object
  detection.
\newblock {\em CVPR}.

\end{thebibliography}

\end{document}